\documentclass[a4paper,fleqn]{cas-sc}

\usepackage[authoryear]{natbib}
\usepackage{lineno}
\usepackage[nameinlink,noabbrev]{cleveref}
\usepackage{chngcntr}
\usepackage{multicol}
\usepackage{nomencl}
\makenomenclature
\usepackage{tabularx}
\crefname{figure}{Fig.}{Figs.}
\crefname{table}{Table}{Tables}
\crefname{equation}{Eq.}{Eqs.}
\Crefname{equation}{Eq.}{Eqs.}

\def\tsc#1{\csdef{#1}{\textsc{\lowercase{#1}}\xspace}}
\tsc{WGM}
\tsc{QE}

\begin{document}
\let\WriteBookmarks\relax
\def\floatpagepagefraction{1}
\def\textpagefraction{.001}

\shorttitle{A Physics Informed Machine Learning Framework for Optimal Sensor Placement and Parameter Estimation}    

\shortauthors{Venianakis et al}  

\title [mode = title]{A Physics Informed  Machine Learning Framework for Optimal Sensor Placement and Parameter Estimation}  


\tnotetext[1]{} 

%

\author[1]{Georgios Venianakis}

\affiliation[1]{organization={School of Chemical Engineering, National Technical University of Athens},
            addressline={Iroon Polytechneiou 9}, 
            city={Athens},
            postcode={15772}, 
            country={Greece}}
\author[1,2]{Constantinos Theodoropoulos}
\affiliation[2]{organization={Department of Chemical Engineering, University of Manchester},
            city={Manchester},
            postcode={M13 9PL}, 
            country={United Kingdom}}
\cormark[1]
\ead{k.theodoropoulos@manchester.ac.uk}

\author[1]{Michail Kavousanakis}
\cormark[1]
\ead{mihkavus@chemeng.ntua.gr}

\cortext[1]{Corresponding authors}


\begin{abstract}
Parameter estimation remains a challenging task across many areas of engineering. 
Because data acquisition can often be costly,  limited, or prone to inaccuracies (noise, uncertainty) it is crucial to identify sensor configurations that provide the maximum amount of information about the unknown parameters, in particular for the case of distributed-parameter systems, where spatial variations are important.
Physics-Informed Neural Networks (PINNs) have recently emerged as a powerful machine-learning (ML) tool for parameter estimation, particularly in cases with sparse or noisy measurements, overcoming some of the limitations of traditional optimization-based and Bayesian approaches.
Despite the widespread use of PINNs for solving inverse problems, relatively little attention has been given to how their performance depends on sensor placement.
This study addresses this gap by introducing a comprehensive PINN-based framework that simultaneously tackles optimal sensor placement and parameter estimation.
Our approach involves training a PINN model in which the parameters of interest are included as additional inputs. 
This enables the efficient computation of sensitivity functions through automatic differentiation, which are then used to determine optimal sensor locations exploiting the D-optimality criterion. 
The framework is validated on two illustrative distributed-parameter reaction-diffusion-advection problems of increasing complexity. 
The results demonstrate that our PINNs-based methodology  consistently achieves higher accuracy compared to parameter values estimated from intuitively or randomly selected sensor positions.

\end{abstract}


\begin{highlights}
\item A PINN framework for joint sensor placement and parameter estimation is presented.
\item Sensor locations are optimized using sensitivity functions and D-optimal design.
\item The framework is demonstrated for reaction-diffusion-advection systems.
\item Optimal placement yields improved parameter estimates over heuristic choices.
\end{highlights}



\maketitle

\section*{Nomenclature}
\begin{tabularx}{\linewidth}{lX} 
NN & Neural Network \\
PINN & Physics Informed Neural Network \\
FIM & Fisher Information Matrix \\
AD & Automatic Differentiation \\
RAR-D & Residual-based Adaptive Refinement with Distribution \\
FEM & Finite Element Method \\
$\mathbf{u}$ & General state variable(s) \\
$\lambda$ & Unknown parameters vector \\
$\lambda^{prior}$ & A priori parameter estimate vector \\
$s_{\lambda_j}$ & Sensitivity function to the parameter $\lambda_j$ \\
$F$ & FIM \\
$\mathbf{x}^*_1, \dots, \mathbf{x}^*_N$ & Selected spatial coordinates of N sensors \\
$\tilde{\cdot}$ & Dimensionless form of a variable \\
$\hat{\mathbf{u}}$ & Approximation of $\tilde{\mathbf{u}}$ (Numerical or PINN) \\
$\theta$ & NN's trainable parameters (weights and biases) \\
$\mathcal{L}(\theta)$ & NN's total loss function \\
$\mathcal{L}_f(\theta)$ & Loss function term corresponding to the PDE residual(s) \\
$\mathcal{L}_{ic}(\theta)$ & Loss function term corresponding to the initial condition residual \\
$\mathcal{L}_{bc}(\theta)$ & Loss function term corresponding to the boundary conditions residuals \\
$\mathcal{L}_{sens}(\theta)$ & Loss function term corresponding to the sum of partial derivatives of all residuals w.r.t. all the parameters of interest \\
$c$ & Species concentration $(kg/m^3)$ \\
$\mathbf{v}$ & Velocity Field $(m/s)$ \\
$p$ & Pressure $(Pa)$ \\
$D$ & Diffusion coefficient $(m^2/s)$ \\
$k$ & Reaction rate constant of an n-th order reaction $((kg/m^3)^{1-n}\cdot s^{-1})$ \\
$U$ & Fluid's mean inlet velocity $(m/s)$ \\
$\rho$ & Fluid density $(kg/m^3)$ \\
$\mu$ & Fluid viscosity $(Pa\cdot s)$ \\
$\tau$ & Characteristic time scale $(s)$ \\
$Pe$ & Péclet number \\
$Da$ & Damköhler number \\
$Re$ & Reynolds number \\
\end{tabularx}

\section{Introduction}\label{introduction}
Distributed-parameter systems, based on partial differential equations (PDEs), typically describing spatiotemporal variations, represent a wide range of physical systems and phenomena in science and engineering.
These PDE-based models typically involve multiple unknown parameters, often corresponding to physical properties of the system, whose accurate identification from available data is essential for reliable prediction and effective system design. 
In general, the accuracy of parameter estimation improves as more data is collected.
Nevertheless, acquiring large datasets is often expensive or impractical.
This motivates the development of systematic methods for selecting sensor locations that maximize the information gained about the unknown parameters under constraints on the number of sensors. 

Most research on Optimal Sensor Placement has focused on state estimation, while comparatively fewer studies have addressed parameter estimation directly \citep{theodoropoulos,theodoropoulos2}. 
Existing approaches typically rely on scalar measures derived from the Fisher Information Matrix (FIM), which captures the sensitivity of observable quantities with respect to parameters.
For example, the modified E-criterion \citep{e_criterion1,e_criterion2,e_criterion3} minimizes the ratio of the largest to the smallest eigenvalue of the FIM, while \cite{trace_inverse} proposed minimizing the expected Bayesian loss involving the trace of the inverse FIM. 
The widely used D-optimality criterion \citep{qureshi,alain} maximizes the determinant of the FIM (or the Gram determinant), and a closely related approach minimizes the information entropy  \citep{entropy}, reflecting the uncertainty of the parameter under a given sensor configuration.
In large-scale distributed systems, parameter estimation can be facilitated by combining model reduction with optimal sensor placement. 
In \cite{alonso2004optimal}, the proposed workflow is to construct a Proper Orthogonal Decomposition (POD) reduced-order model that captures the dominant spatio-temporal dynamics and then determine measurement locations by optimizing an observability-based criterion on this reduced model. 
\cite{theodoropoulos} further streamline this process by proposing sensor placement directly at the extrema of the dominant POD modes, an approach that circumvents costly sensitivity-matrix computations while still providing near-optimal parameter estimation performance.
Once sensor locations are defined, the next step is parameter estimation.
Optimization-based methods, such as Maximum Likelihood Estimation (MLE) \citep{MLE}, remain the most established, with Least Squares as the most common implementation under the assumption of Gaussian measurement noise \citep{mle1,mle2}. 
While effective, these methods typically require repeated model evaluations, resulting in high computational cost, particularly when gradients with respect to parameters must also be computed.
Moreover, the accuracy of estimated parameters may degrade under sparse or noisy data \citep{mle_noise}. 
Bayesian methods offer an alternative by incorporating prior information and providing full uncertainty quantification through posterior distributions \citep{bayesian3,bayesian1,bayesian2}. 
Nevertheless, Bayesian inference often relies on Markov Chain Monte Carlo (MCMC) sampling \citep{mcmc}, which requires extensive forward model evaluations and can be sensitive to prior selection in data-limited settings \citep{bayes_priors}.
Recent advances in machine learning have substantially enhanced the field of parameter estimation by introducing data-driven methodologies capable of capturing complex relationships between model parameters and observed data. 
Traditional statistical approaches, such as maximum likelihood estimation and Bayesian inference, have been extended through modern techniques, such as variational inference \citep{blei2017variational} and neural density estimation \citep{papamakarios2021normalizing} that enable efficient and accurate parameter recovery, even in complex or high-dimensional settings. 
Furthermore, deep neural architectures, coupled with gradient-based optimization algorithms, including stochastic gradient descent and its adaptive variants \citep{bottou2010large}, facilitate scalable and robust inference across a wide range of applications.
Hybrid modeling frameworks that integrate data with first-principles formulations have merged as powerful tools for scientific computing.  
Among these, Physics-Informed Neural Networks (PINNs) \citep{pinns_og} exploit the expressivity of deep neural networks \citep{universal_aprox} while embedding governing PDEs as soft constraints. 
PINNs have proven particularly useful in scenarios where traditional solvers face challenges: complex geometries (since PINNs are mesh-free), high-dimensional problems \citep{curse_of_dim}.  and systems with incomplete physics that require data-driven enhancement \citep{incomplete_model}.
Importantly, PINNs enable parameter estimation by directly incorporating unknown parameters into the training process, making them robust even under sparse or noisy observations, which is of particular relevance to this work. 
Numerous studies have applied PINNs to inverse problems in chemical engineering and related fields. 
Examples include the estimation of thermal properties in additive manufacturing \citep{additive}, thermal diffusivity in two-phase flows \citep{heat_transfer}, kinetic parameters in catalytic reactors \citep{co2_reactor, membrane_reactor}, and transport coefficients in porous media \citep{porous}.
PINNs have also been used in fluid mechanics, such as inferring density fields in high-speed flows \citep{pinns_high_speed} and thrombus properties in arterial flows \citep{pinn_thrombus}.
Despite these successes, most studies assume dense observations across the spatial domain, often generated from numerical simulations, overlooking the constraints of real-world sensor settings. 
For instance, \cite{co2_reactor} observed that parameter estimation accuracy strongly depends on the spatial region from which data is collected, but this conclusion was based on manual experimentation rather than systematic sensor placement strategies. 
The integration of optimal sensor placement with PINNs remains relatively unexplored.
A few attempts have been made: \cite{gs_pinn} proposed a greedy sampling approach (GS-PINN), which selects informative samples via Proper Orthogonal Decomposition (POD) and the Discrete Empirical Interpolation (DEIM) method \citep{deim}, but their approach requires access to full spatio-temporal data and selects time-varying sensor positions, limiting practical use. 
In \cite{sbs_pinn}, they introduce a sensitivity-based sampling method (SBS) that adapts sensor placement based on gradients of the PINN loss with respect to collocation points. 
However, this method was designed for state reconstruction problems in process control, not parameter estimation, as it does not account for sensitivities with respect to the parameters.
In this work, we address this gap by proposing a PINN framework that integrates D-optimal sensor placement directly into the parameter estimation process. 
Our approach relies on two complementary neural network models. 
The first model incorporates the parameters of interest as inputs, enabling the computation of sensitivity functions via automatic differentiation (AD).
These sensitivities are used to select optimal sensor locations under the D-optimality criterion. 
The second, is a standard PINN model trained on data obtained from these optimal locations to perform parameter estimation and state reconstruction. 

We demonstrate our method on two illustrative case studies involving reaction-diffusion-advection systems of increased complexity: a one-dimensional steady-state problem, and a two-dimensional transient problem. In each case, the goal is to estimate important dimensionless parameters for the system such as the P\'{e}clet number and Damk\"{o}hler number and to compare their accuracy against estimations made from sub-optimal sensor locations.

The paper is organised as follows: \Cref{methods} presents the formulation of our method, including sensor selection and PINN training. 
\Cref{case study} describes the case studies and results.
Finally, \Cref{conclusions} discusses the main conclusions and the relevant future directions.
\section{Methods}\label{methods}

\subsection{Selection of Optimal Sensor Location for Parameter Estimation}\label{opt_sensor}

In this work, we adopt the D-optimality criterion, widely used for sensor placement due to its geometric interpretability and invariance to parameter rescaling \citep{d_optimal_choice1,d_optimal_choice2}. 
For implementation, we follow the method of \citet{alain}, chosen for its simplicity.
Alternative optimality criteria can also be applied as discussed in \citep{theodoropoulos}.

Consider a system with states $\mathbf{u}(\mathbf{x},t)$ described by the governing PDE: 
\begin{equation}
    \frac{\partial\mathbf{u}}{\partial t} + \mathcal{N}[\mathbf{u};\lambda] = 0,\quad \mathbf{x} \in \Omega,\quad t \ge 0,
    \label{eq: pde form}
\end{equation}
\noindent subject to the following initial and boundary conditions:
\begin{align}
    &\mathbf{u}(0,\mathbf{x}) = \mathbf{g}(\mathbf{x}), \quad \mathbf{x} \in \Omega, \label{eq: initial cond} \\[4pt]
    &\mathcal{B}[\mathbf{u}] = 0, \quad \mathbf{x} \in \partial \Omega,\quad t \ge 0, \label{eq: bound cond}
\end{align}

\noindent where $\mathcal{N}[\cdot]$ is a differential operator, and $\mathcal{B}[\cdot]$ a boundary operator, $\mathbf{\lambda}=\left[\lambda_1, \dots,\lambda_P  \right]$ are the $p$ unknown parameters, and $\Omega$ the spatial domain.

For state measurements at $N$ sensor locations $\mathbf{x}_1, \dots, \mathbf{x}_N$ we define sensitivity functions as:
\begin{equation}
    s_{\lambda_j}(\mathbf{x}_i,t) = \frac
    {\partial \mathbf{u}}
    {\partial \lambda_j}(\mathbf{x}_i,t), \;
    i = 1,\dots,N,\; j = 1,\dots,P,
    \label{sensitivity functions}
\end{equation}

\noindent and collect them into a vector $M$:
\begin{equation}
\begin{aligned}
M(x_1, \ldots, x_N, t) = \big[ 
s_{\lambda_1}(\mathbf{x}_1, t) \;\ldots\; s_{\lambda_P}(\mathbf{x}_1, t) \;\ldots 
s_{\lambda_1}(\mathbf{x}_N, t) \;\ldots\; s_{\lambda_P}(\mathbf{x}_N, t) 
\big]^T.
\end{aligned}
\label{eq:M_vector}
\end{equation}
The Fisher Information Matrix, $F$, is then:

\begin{equation}
    F(\mathbf{x}_1, \dots, \mathbf{x}_N) = \int_0^T M(x_1, \ldots, x_N, t)M^T(x_1, \ldots, x_N, t)\,dt,
    \label{eq: FIM}
\end{equation}

\noindent where $[0,T]$ is the observation horizon.
The matrix $F$ is positive semi-definite with its diagonal entries quantifying sensor sensitivity to parameters, while its off-diagonal entries capture correlations between the corresponding parameters.

The optimal sensor locations $\mathbf{x}^*_1, \dots, \mathbf{x}^*_N$ maximize the Gram determinant \citep{gramdet}:
\begin{equation}
    \mathbf{x}^*_1, \dots, \mathbf{x}^*_N = \arg\max_{\mathbf{x}_1, \dots, \mathbf{x}_N} \det\left[F(\mathbf{x}_1, \dots, \mathbf{x}_N) \right].
\end{equation}

Maximizing $\det(F)$ encourages both large sensitivities (via diagonal dominance) and independence (via small off-diagonal terms).
For \textit{steady-state} problems the FIM reduces to:
\begin{equation}
    F(\mathbf{x}_1, \dots, \mathbf{x}_N) = M(x_1, \ldots, x_N, t)M^T(x_1, \ldots, x_N, t),
\end{equation}

\noindent which is rank-1 with zero determinant, rendering D-optimality unusable.
For these cases, we instead select optimal sensor locations by maximizing the FIM trace:
\begin{equation}
    \mathbf{x}^*_1, \dots, \mathbf{x}^*_N = \arg\max_{\mathbf{x}_1, \dots, \mathbf{x}_N} \operatorname{tr}\left[F(\mathbf{x}_1, \dots, \mathbf{x}_N) \right].
\end{equation}

Our PINN framework as well as the calculation of sensitivities using PINNs is discussed next.

\subsection{Physics Informed Neural Networks: General Framework}\label{pinns}

PINNs are fully connected neural networks that incorporate PDE constraints into their loss function (\citet{pinnframework}).
They address two problem types:
\begin{enumerate}[(a)]
    \item \textbf{Forward Problems:} approximating PDE solutions with known parameters and boundary/initial conditions, without experimental data. 
    \item \textbf{Inverse Problems:} estimating unknown parameters from experimental data, while simultaneously approximating the PDE solution.
\end{enumerate}
For the forward problem, the system state $\mathbf{u}(t,\mathbf{x})$ governed by \crefrange{eq: pde form}{eq: bound cond} is approximated by a neural network $\hat{\mathbf{u}}(t,\mathbf{x},  \mathbf{\theta})$, where $\mathbf{\theta}$ denotes trainable weights and biases.
Training minimizes a composite loss function, $\mathcal{L}$:

\begin{equation}
    \mathcal{L}(\mathbf{\theta}) = \mathcal{L}_{f}(\mathbf{\theta}) +\mathcal{L}_{ic}(\mathbf{\theta}) + \mathcal{L}_{bc}(\mathbf{\theta}),
\end{equation}

\noindent with terms enforcing PDE residuals ($\mathcal{L}_{f}$), initial conditions ($\mathcal{L}_{ic}$), and boundary conditions ($\mathcal{L}_{bc}$).
Automatic differentiation computes derivatives of the system state required for residual evaluation (PDE and boundary conditions residuals).
For \textit{inverse problems}, observational data $\left\{\mathbf{u}_{data}^i \right\}_{i=1}^{N_{data}}$ at spatio-temporal coordinates $\left\{ t_{data}^{i}, \mathbf{x}_{data}^{i} \right\}_{i=1}^{N_{data}}$ introduce an additional loss term:

\begin{equation}
    \mathcal{L}_{data}(\mathbf{\theta}) = \frac{1}{N_{data}} \sum_{i=1}^{N_{data}} \left| \hat{\mathbf{u}}(t_{data}^{i}, \mathbf{x}_{data}^{i}) - \mathbf{u}_{data}^i \right|^2,
    \label{eq: data loss}
\end{equation}
\noindent leading to the total loss function:

\begin{equation}
    \mathcal{L}(\mathbf{\theta}) = \mathcal{L}_{ic}(\mathbf{\theta}) + \mathcal{L}_{bc}(\mathbf{\theta}) + \mathcal{L}_{f}(\mathbf{\theta}) + \mathcal{L}_{data}(\mathbf{\theta}).
    \label{eq: general PINN loss function}
\end{equation}

In this case, PINNs are optimized with respect to both $\boldsymbol{\theta}$ and unknown parameters $\boldsymbol{\lambda}$.

\subsection{Sensitivity functions with PINNs}\label{sensitivity}
As described in \Cref{opt_sensor}, optimal sensor placement requires sensitivity functions (\Cref{sensitivity functions}).
Following \citet{sensitivity_PINNs}, the PINN input is extended to include both spatio-temporal coordinates and parameters $\lambda$, enabling AD-based computation of $\partial{\hat {\mathbf{u}}}/\partial{\lambda_j}$.
Additional loss terms enforce accurate parameter sensitivities calculation:

\begin{equation}
\begin{split}
    \mathcal{L}(\mathbf{\theta}) = \mathcal{L}_{ic}(\mathbf{\theta}) + \mathcal{L}_{bc}(\mathbf{\theta}) + \mathcal{L}_{f}(\mathbf{\theta}) 
    + \mathcal{L}_{sens,ic}(\mathbf{\theta})
    + \mathcal{L}_{sens,bc}(\mathbf{\theta}) + \mathcal{L}_{sens,f}(\mathbf{\theta}),
\end{split}
\end{equation}

\noindent where:

\begin{equation}
    \mathcal{L}_{sens,ic}(\mathbf{\theta}) =
     \frac{1}{N_{ic}} \sum_{j=1}^P\sum_{i=1}^{N_{ic}}
     \left |\frac{\partial r_{ic}}{\partial \lambda_j}
     (0,\mathbf{x}_{ic}^{i}, \lambda, \mathbf{\theta})
     \right|^2,
\end{equation}
\begin{equation}
    \mathcal{L}_{sens,bc}(\mathbf{\theta}) =
     \frac{1}{N_{bc}} \sum_{j=1}^P\sum_{i=1}^{N_{bc}}
     \left |\frac{\partial r_{bc}}{\partial \lambda_j}
     (t_{bc}^{i}, \mathbf{x}_{bc}^{i}, \lambda, \mathbf{\theta})
     \right|^2,
\end{equation}
\begin{equation}
    \mathcal{L}_{sens,f}(\mathbf{\theta}) =
     \frac{1}{N_{f}} \sum_{j=1}^P\sum_{i=1}^{N_{f}}
     \left |\frac{\partial r_f}{\partial \lambda_j}
     (t_{f}^{i}, \mathbf{x}_{f}^{i}, \lambda, \mathbf{\theta})
     \right|^2.
\end{equation}

\noindent Each $\mathcal{L}_{sens,\cdot}$ term penalizes derivatives of residuals with respect to $\lambda$.

\begin{figure*}
    \centering
    \includegraphics[width=0.9\linewidth]{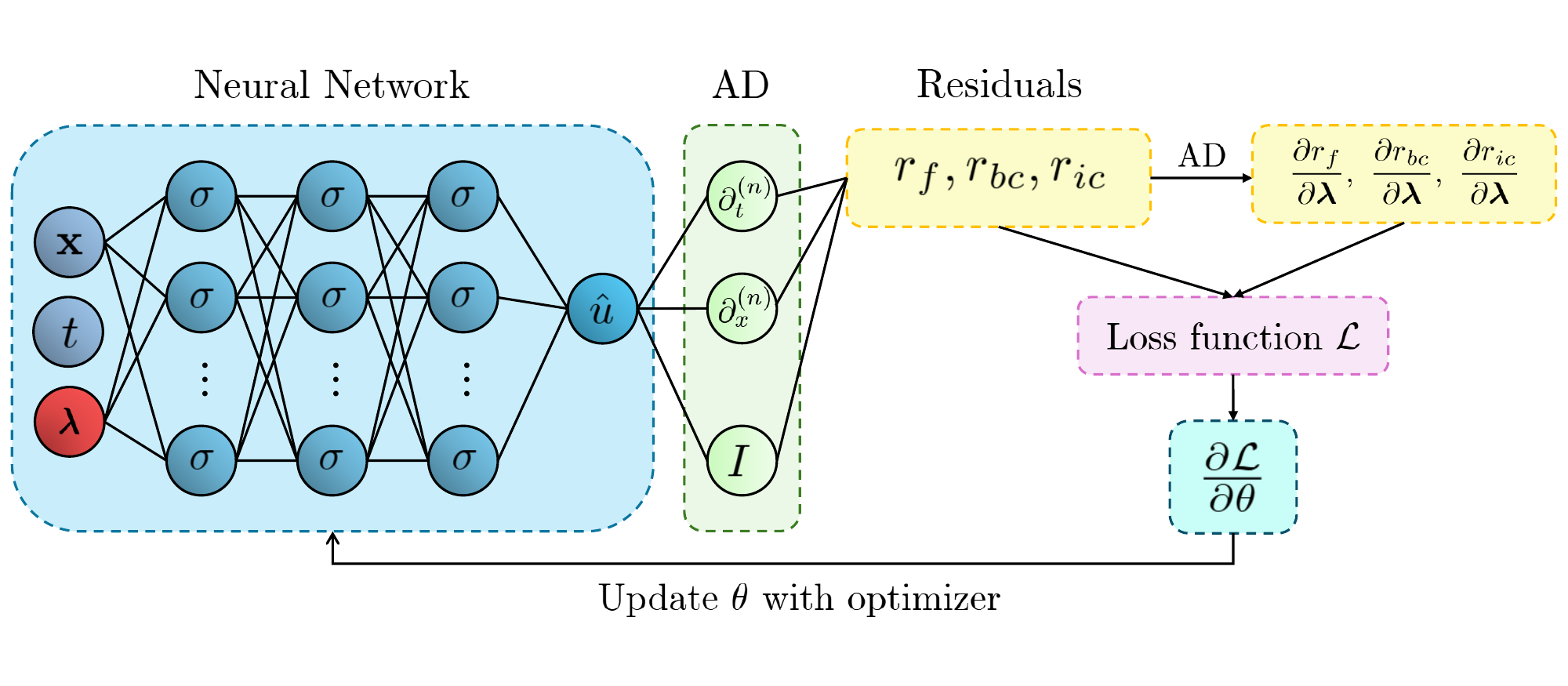}
    \caption{Schematic representation of the methodology for computing sensitivity derivatives using PINNs. 
    The neural network admits as input the spatio-temporal coordinates $(\mathbf{x},t)$ along with the parameters of interest, $\lambda$.
    The loss function $\mathcal{L}$ to be minimized is constructed from the system residuals (PDE and boundary/initial conditions) and their derivatives with respect to the parameters,  $\lambda$. }
    \label{fig: sa-pinn schematic}
\end{figure*}

Training is performed with prior parameter estimates, ${\lambda}^{prior}$.
The resulting sensitivity functions are then used to construct the FIM and determine optimal sensor locations.
Here we assume that their qualitative structure (e.g., locations of maxima) remains the same when evaluated at $\lambda^{prior}$ compared to the true $\lambda$.
When the computed optima differ substantially, we use an iterative approach.
\Cref{fig: sa-pinn schematic} illustrates the process.

Once sensor placement is decided, state data are collected from the corresponding locations, and the inverse problem is solved using PINNs.
To accelerate training, we apply transfer learning \citep{transfer}: the inverse PINN is initialized from a pretrained forward PINN at prior parameter values, $\lambda^{prior}$.
A lightweight fine-tuning strategy is adopted here \citep{transfer_light}, which retrains only the last hidden layers while freezing the earlier ones, reducing both the number of epochs and the per-iteration cost.
Finally, to improve PINN training efficiency and stability we incorporate techniques including:

\begin{itemize}
	\item{\textbf{Dynamic loss weighting:} Following \cite{dynamic_weight}, loss weights $w_k$ multiplying each loss term in the loss function are adaptively updated to balance gradients across different terms, preventing dominance by any single component (see e.g., \Cref{eq: 1d loss function} and \Cref{eq: total loss function fluid} below).}
	\item{\textbf{Residual-based adaptive refinement (RAR-D):} Collocation points are adaptively chosen  according to the PDE residual \citep{rar_d}, focusing training on regions with higher errors.
	We extend this strategy to boundary and initial condition sampling (collocation)  points as well. } 
	\item{\textbf{Non-dimensionalization:} PDEs are solved in dimensionless form, preventing vanishing/exploding gradients and ensuring balanced loss contributions across variables \citep{pinnframework}.}
\end{itemize}
\section{Case Studies}\label{case study}
This section presents two illustrative examples that demonstate our PINNs-based methodology and also showcase how the optimal sensor placement criterion enhances parameter inference accuracy.
All PINN models are implemented in Python using the PyTorch\textsuperscript{\textregistered} library \citep{pytorch} and trained on an NVIDIA\textsuperscript{\textregistered} RTX A4000 GPU computer.

\subsection{1D Steady State Reaction–Advection–Diffusion equation}
\subsubsection{Problem Set-Up}
The first simpler example considers a steady state, one-dimensional reaction-advection-diffusion equation, which in dimensionless form is given by the following PDE and boundary conditions:
\begin{equation}
\begin{split}
    Pe \frac{dc}{dx} = \frac{d^2c}{dx^2} - Da c^2,\; {x} \in \left[0,10 \right],\\
    c(0) = 1, \\
    \left.\frac{dc}{dx}\right|_{x = L} = 0,
\end{split}
\label{PDE1}
\end{equation}
where $Pe$ and $Da$ denote the P{\'e}clet number and the Damk{\"o}hler number, respectively.
In this example, $Da$ is fixed ($Da = 1.0$), and the goal is to estimate the unknown value of $Pe$ with a single, optimally placed sensor.

\subsubsection{PINN Framework}
To compute sensitivities for the FIM construction, we employ a Sensitivity PINN model with two inputs ($x$ and $Pe$), four hidden layers of 30 neurons each, and hyperbolic tangent activation.
This network structure is chosen to balance computational efficiency with accuracy. 
The loss function is
\begin{equation}
\begin{split}
    \mathcal{L}(\theta) = w_f \mathcal{L}_f(\theta) 
    +w_{bc1} \mathcal{L}_{bc1}(\theta)
    +w_{bc2} \mathcal{L}_{bc2}(\theta)
    +w_{sens} 
    \left(\mathcal{L}_{sens,f}(\theta)
    +\mathcal{L}_{sens,bc1}(\theta)
    +\mathcal{L}_{sens,bc2}(\theta) \right),  
\end{split}
\label{eq: 1d loss function}
\end{equation}
\noindent where $\mathcal{L}_f(\theta) $, $\mathcal{L}_{bc1}(\theta)$, $\mathcal{L}_{bc2}(\theta)$ correspond to the PDE and boundary condition residuals. 
The terms $\mathcal{L}_{sens,f}(\theta)$, $\mathcal{L}_{sens,bc1}(\theta)$ and $\mathcal{L}_{sens,bc2}(\theta)$ represent the corresponding derivatives with respect to $Pe$. 
Analytical expressions are provided in Section~I of the Supplementary Material.
We sample 800 collocation points on an equispaced grid over the spatial domain providing sufficient resolution for our PINN model.
The loss weights are dynamically updated following the gradient ascent scheme of \citet{dynamic_weight}.
Training is performed with the Adam optimizer ($10^{-3}$ learning rate) for 2500 epochs. 
Since the true value of $Pe$ is to be estimated (unknown), the model is initialized with an a priori estimate, $Pe = 0.1$.
As noted in \Cref{sensitivity}, we assume that the maxima of the Gram determinant (or the FIM trace in this example) are approximately invariant to whether they are computed at the estimated or true parameter values. 
This is equivalent to having a "reasonable" prior estimate. An iterative procedure can be employed if the computed optima vary significantly.  
%
%
For a single-sensor placement problem, the FIM trace reduces to the squared sensitivity of $c$ with respect to $Pe$.
\begin{equation}
    tr\left[F(x) \right] = \left[ \frac{\partial c}{\partial Pe}(x) \right]^2,\; {x} \in \left[0,10 \right].
\end{equation}

\begin{figure}[ht]
    \centering
    \includegraphics[width=0.55\linewidth]{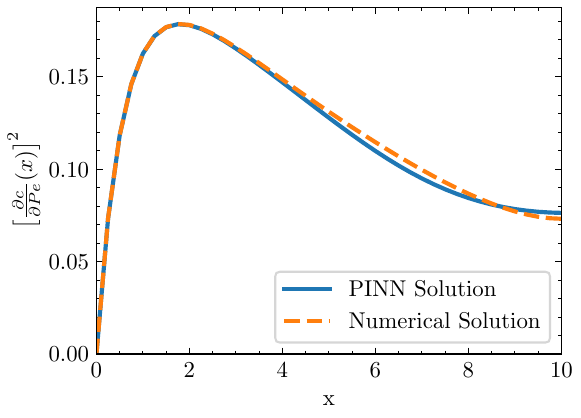}
    \caption{Spatial profile of the FIM trace computed by PINNs and numerically for $Pe = 0.1$ ({\it a priori} estimate) and $Da = 1.0$.}
    \label{fig: 1d trace}
\end{figure}

\Cref{fig: 1d trace} shows the FIM trace obtained from the PINN alongside the one obtained by numerically soving the PDE (equation \ref{PDE1})
The results confirm that PINNs accurately compute sensitivity derivatives via automatic differentiation. 
Here, it is easy to locate graphically the trace maximum at $x^* = 1.81$. 

Consequently, pseudo-experimental data are  generated numerically using the true value of $Pe = 1.0$ at two locations:
the optimal location $x^* = 1.81$, and a suboptimal point at the outlet, $x^* = 10.0$, which is an intuitive choice for placing the sensor.
The inference PINN has the same architecture but accepts only $x$ as input, with $Pe$ treated as a trainable parameter.
The loss function now reads:

\begin{equation}
\begin{split}
    \mathcal{L}(\theta) = w_f \mathcal{L}_f(\theta) 
    +w_{bc1} \mathcal{L}_{bc1}(\theta) 
    +w_{bc2} \mathcal{L}_{bc2}(\theta) + w_{data} \mathcal{L}_{data}(\theta),
\end{split}
\end{equation}

\noindent with the data term being:

\begin{equation}
    \mathcal{L}_{data}(\mathbf{\theta}) = \frac{1}{N_{data}} \sum_{i=1}^{N_{data}} \left| \hat{c}({x}_{data}^{i}) - c_{data}^i \right|^2.
\end{equation}
Since a single sensor is used, $N_{data} = 1$.
The model is trained with the Adam optimizer for 5000 iterations.
The hyperparameters for the Sensitivity and Inference PINNs applied to the 1D steady-state reaction-advection-diffusion equation are summarized in \Cref{tab:pinn_hyperparameters}.

\begin{figure}[ht]
    \centering
    \includegraphics[width=0.55\linewidth]{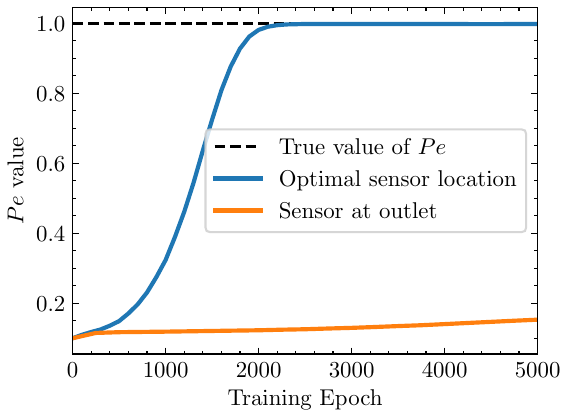}
    \caption{Inference of $Pe$ vs. PINN training epochs with sensor placed at the FIM trace maximum ($x^* = 1.81$ -blue curve) versus at the outlet ($x^* = 10.0$ -red curve)}
    \label{fig: 1d Pe inf}
\end{figure}

\Cref{fig: 1d Pe inf} shows the inferred $Pe$ during training for data from the optimal and outlet locations.
The results highlight the crucial role of sensor placement: the PINN converges closer to the true $Pe$,  when data are taken at the optimal location. 
At the outlet, the parameter fails to converge for the number of epochs used.
Final estimates and relative errors are listed in \Cref{results case 1}.

\begin{table*}[ht]
\centering
\caption{Hyperparameters of Sensitivity PINNS and Inference PINNs.}
\begin{tabular}{@{}lll@{}}
\toprule
\textbf{PINN Hyperparameter}         & \textbf{Sensitivity PINN} & \textbf{Inference PINN} \\ \midrule
Hidden layers               & 4                         & 4                     \\
Neurons per layer           & 30                        & 30                    \\
Inputs                      & 2 ($x$, $Pe$)                 & 1 ($x$)                 \\
Activation function                   & \texttt{$\textrm{tanh}()$}           & \texttt{$\textrm{tanh}()$}       \\
Optimizer                             & Adam                      & Adam                  \\
Iterations                  & 2500                      & 5000\\
Collocation points          & 800                       & 800                   \\
Dynamic weight strategy               & Yes                       & Yes                   \\
RAR-D                                 & No                        & No                    \\ \bottomrule
\end{tabular}
\label{tab:pinn_hyperparameters}
\end{table*}

\begin{table}[h!]
\centering
\caption{Estimated $Pe$ values and relative error for optimal versus suboptimal sensor placement.}
\label{results case 1}
\begin{tabular}{lll} 
\toprule
\textbf{Sensor} & \textbf{Estimated $Pe$ (true: 1.000)} & \textbf{Relative Error (\%)} \\
\midrule
Optimal ($x^* = 1.81$) & 0.998 & 0.20 \\
Outlet ($x^* = 10.0$) & 0.153 & 84.7 \\
\bottomrule
\end{tabular}
\end{table}

\begin{figure}[h]
    \centering
    \includegraphics[width=0.55\linewidth]{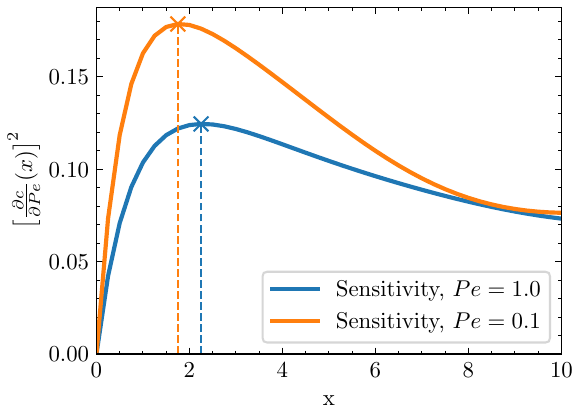}
    \caption{FIM trace for the {\it a priori} estimate $Pe=0.1$ and true value $Pe=1.0$.}
    \label{fig: 1d sens init vs true}
\end{figure}

After identifying the parameter, we recompute the optimal sensor location using the true $Pe$.
As shown in \Cref{fig: 1d sens init vs true}, the maximum shifts slightly to $x^* = 2.32$, very close to the original $x^* = 1.81$ supporting the assumption that sensitivity maxima remain nearly invariant when evaluated with initial parameter estimates.

\subsection{2D Transient Reaction–Advection–Diffusion, Flow Around a Fixed Obstacle}

The 1D example demonstrates how our PINN-based methodology guides sensor placement and enables accurate estimation of a single parameter. 
To further test the robustness of the framework, we now consider a more complex 2D system that combines nonlinear reaction, advection and diffusion in the presence of an obstacle. 
This example introduces additional challenges: heterogeneity due to flow recirculation, transient behavior, as well as the simultaneous inference of two parameters. 
In particular, we address the simultaneous estimation of the dimensionless parameters $Pe$ and $Da$ in a two-dimensional, time-dependent reaction-advection-diffusion system. 
A reactive species is transported by a steady flow past a fixed cylindrical obstacle.

\subsubsection{Problem Set-Up}
We consider a 2D rectangular channel of dimensions $L \times H$ ($L = 0.24$ m and $H = 0.14$ m), with flow from left to right bounded by impermeable walls at the top and bottom.
A fixed cylindrical obstacle of radius $R = 0.01$ m is placed inside the channel perpendicular to the flow direction, at $\left \{x_0,y_0 \right\} =\left \{0.09, 0.07 \right\}$ m.
The fluid is incompressible, Newtonian, and the flow is considered to be at  steady-state.
Its velocity, ${\mathbf{v}}$, and pressure, $p$, fields satisfy the Navier-Stokes equations (neglecting gravity):
\begin{equation}
    \rho \left (\frac{\partial \mathbf{v}}{\partial t}+\mathbf{v\cdot\nabla\mathbf{v}}    \right) =  -\nabla p + \mu \nabla^2\mathbf{v},
    \label{eq: navier-stokes}
\end{equation}

\noindent subject to the incompressibility constraint: 

\begin{equation}
    \nabla\cdot\mathbf{v} = 0.
    \label{eq: continuity}
\end{equation}

\noindent where $\rho, \mu$ denote the density and dynamic viscosity, respectively.

At the inlet, a parabolic velocity profile is imposed for the horizontal component of velocity, $v_x$: 

\begin{equation}
    v_x(0,y) = 6 U \left[ \left(\frac{y}{H}\right) - \left(\frac{y}{H}\right)^2 \right].
    \label{eq:inlet velocity profile}
\end{equation}

\noindent with no-slip conditions at the walls (including the obstacle surface) and zero pressure at the outlet.
The fluid properties are set to $\rho = 1000$ $\left[{kg}/{m^3}\right]$,  $\mu = 0.001$ $\left [Pa \cdot s \right]$ and average inlet velocity $U = 0.00125$ $\left[{m}/{s} \right]$.

\begin{figure*}[ht]
    \centering
    \includegraphics[width=1.0\linewidth]{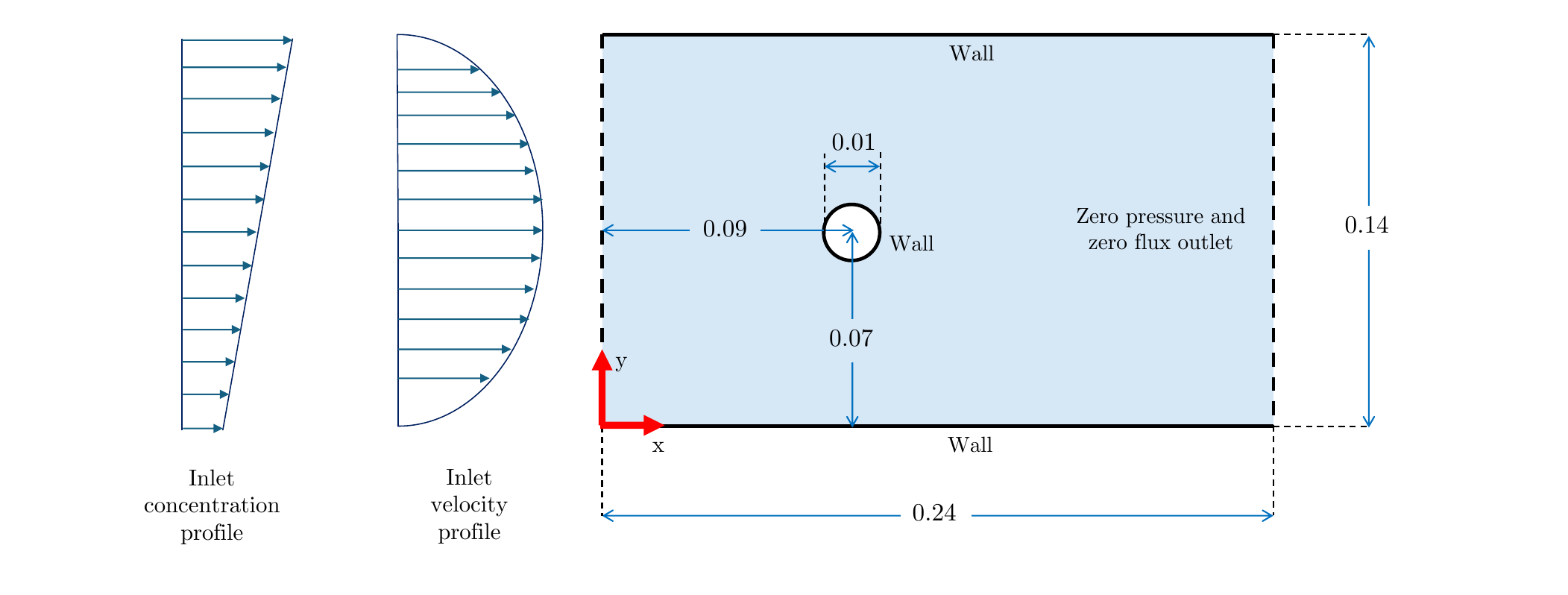}
    \caption{Computational domain: the fluid enters the channel with a parabolic velocity profile at the horizontal direction carrying a reactive species, flows past a fixed obstacle, and undergoes transport and reaction.}
    \label{fig: comp domain}
\end{figure*}

Given the steady flow $\mathbf{v}$, the concentration $c(x,y,t)$ of a reactive species evolves according to:
\begin{equation}
\frac{\partial c}{\partial t} + \mathbf{v} \cdot \nabla c = D \nabla^2 c - k c^2,  \ (x, y) \in [0, L] \times [0, H], \ t>0,
\label{eq:general_mass_equation}
\end{equation}

\noindent where $D$ is the diffusion coefficient and $k$ the reaction rate constant. 
The initial condition is:
\begin{equation}
    c(x,y,0) = 0,  \ (x, y) \in (0, L] \times [0, H].
\end{equation}

\noindent The inlet concentration profile is set to change linearly with $y$:

\begin{equation}
    c(0,y,t) = \frac{y}{H} + 0.3 \left[\frac{kg}{m^3} \right],\ y \in[0,H],\ t>0.
\end{equation}

Zero-flux boundary conditions are applied at the impenetrable walls and the (long) outlet.
A schematic of the setup is shown in \Cref{fig: comp domain}.
By selecting as characteristic scales, $L$, $U$, and $c_0=1$ kg/m$^3$, we derive the following dimensionless variables:
\[\tilde{x} = \frac{x}{L},\ \tilde{y} = \frac{y}{L}, 
\ \tilde{\mathbf{v}} = \frac{\mathbf{v}}{U},\ \tilde{p} = \frac{p}{\rho {U}^2} \   \text{and} \ \tilde{c} = \frac{c}{c_0}.\]

Typically, time is scaled using 
\[
\tilde{t} = \frac{tD}{L^2}.
\]
\noindent However, in this problem, the diffusion coefficient $D$ is unknown and will be determined implicitly through the dimensionless parameters. 
Therefore, it cannot serve as the basis for non-dimensionalizing time. 
Instead, we define a characteristic time scale, $\tau$, chosen to be of the same order of magnitude as ${L^2}/{D_{0}}$, where $D_0$ is an initial estimate of the diffusion coefficient.
Numerical tests show that setting $\tau = 4000 $ s yields good results.
The dimensionless time variable is therefore defined as:
\[
\tilde{t} = \frac{t}{\tau}.
\]
Thus, in dimensionless form, the governing equations are:

\begin{equation}
    \tilde{\mathbf{v}} \cdot \nabla_{\tilde{\mathbf{x}}} \tilde{\mathbf{v}} = -\nabla_{\tilde{\mathbf{x}}} \tilde{p} + \frac{1}{Re} \nabla^2_{\tilde{\mathbf{x}}}\tilde{\mathbf{v}},
    \label{eq: nav-stokes nondim}
\end{equation}

\begin{equation}
    \nabla_{\tilde{\mathbf{x}}} \cdot     \tilde{\mathbf{v}}=0,
    \label{eq: continuity nondim}
\end{equation}

\begin{equation}
\begin{split}
    Pe\frac{L}{\tau U} \frac{\partial \tilde{c}}{\partial \tilde{t}} + Pe  \tilde{\mathbf{v}} \cdot \nabla_{\tilde{\mathbf{x}}} \tilde{c}
    = 
    \nabla^2_{\tilde{\mathbf{x}}}{\tilde{c}} - Da \ {\tilde{c}^2}, \quad 
\tilde{\mathbf{x}} \in [0, 1] \times [0, \frac{H}{L}], \quad \tilde{t} \ge 0, 
\end{split}
\label{eq: mass non_dim}
\end{equation}
where $Re = \frac{\rho U L}{\mu}$, $Pe = \frac{UL}{D}$ and $Da = \frac{k c_0L^2}{D}$ are the Reynolds, P{\'e}clet and Damk{\"o}hler numbers, respectively.
From this point forward, tildes are omitted for convenience.

\subsubsection{PINN Set-Up}
Following the approach of \cite{multi_physics}, we construct and train separate PINN models: 
one for the fluid flow PDEs (Navier-Stokes, \Cref{eq: nav-stokes nondim,eq: continuity nondim}) and another for the species transport PDE (\Cref{eq: mass non_dim}).
Because of the one-way coupling (fluid flow influences mass transport but not vice versa for a dilute system) and since no flow-field parameter inference is required, we first train the fluid flow PINN using the fixed parameters and boundary conditions described earlier.
Its output is then used to compute the residual of \Cref{eq: mass non_dim}, which in turn guides the training of two distinct mass transport PINNs:
one for computing sensitivity derivatives and another for parameter estimation.

\subsubsection*{Fluid Flow PINN}
\begin{figure}[h]
    \centering
    \includegraphics[width=1\linewidth]{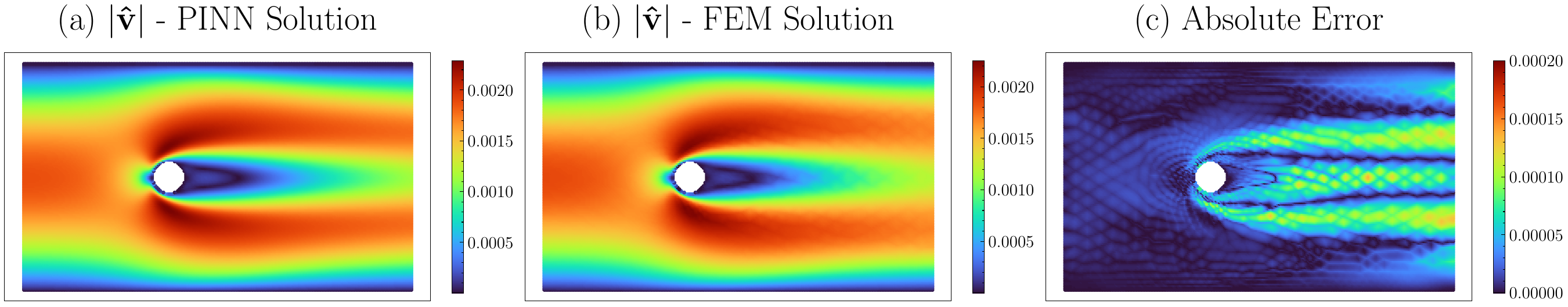}
    \caption{Magnitude of the velocity field $|\mathbf{\hat{v}}|$: (a) PINN prediction of the fluid flow, (b) FEM numerical solution and (c) absolute error of PINN prediction relative to the FEM solution. }
    \label{fig: flow_field}
\end{figure}

The fluid flow PINN admits two inputs, $x$ and $y$, and produces: $\hat{\mathbf{v}}$, and $\hat{p}$.
The $\ \hat{}\ $ symbol is used to denote the PINN approximation.
The architecture is a fully connected neural network with 8 hidden layers of 60 neurons each, using hyperbolic tangent activation functions.
The loss function minimized during training is:
\begin{equation}
\begin{split}
    \mathcal{L}(\mathbf{\theta}) &= w_{in} \mathcal{L}_{in}(\mathbf{\theta}) 
    + w_{out} \mathcal{L}_{out}(\mathbf{\theta}) 
     + w_{wall,up} \mathcal{L}_{wall,up}(\mathbf{\theta}) 
    + w_{wall, low} \mathcal{L}_{wall, low}(\mathbf{\theta}) \\
    &\quad + w_{wall,obst.} \mathcal{L}_{wall,obst.}(\mathbf{\theta}) 
    + w_{ns} \mathcal{L}_{ns}(\mathbf{\theta})
    + w_{cont.} \mathcal{L}_{cont.}(\mathbf{\theta})   ,
\end{split}
\label{eq: total loss function fluid}
\end{equation}

\noindent where $\mathcal{L}_{ns}(\mathbf{\theta})$ and $\mathcal{L}_{cont.}(\mathbf{\theta})$ correspond to the residuals of \Cref{eq: nav-stokes nondim,eq: continuity nondim}, and $\mathcal{L}_{in}(\mathbf{\theta})$, $\mathcal{L}_{out}(\mathbf{\theta})$, $\mathcal{L}_{wall,up}(\mathbf{\theta})$, $\mathcal{L}_{wall,low}$, and $\mathcal{L}_{wall,obst.}$ correspond to the inlet, outlet and wall boundary condition residuals, respectively. 
Analytical expressions for each loss term are provided in Section~II.1 of the Supplementary Material.
The PINN model is trained for 35,000 iterations with the Adam optimizer, followed by 2,000 iterations with L-BFGS for improved accuracy.
We also employ RAR-D (adaptive collocation point sampling) and an  adaptive weight strategy (see \Cref{sensitivity}).
Complete hyperparameter settings are listed in \Cref{tab:pinn_hyperparams_all}.
Training requires approximately 95 minutes. 
The magnitude of the predicted flow field, $|\mathbf{\hat{v}}| = \sqrt{\hat{v}_x^2 + \hat{v}_y^2}$, is visualized and compared with the corresponding numerical solution in \cref{fig: flow_field} showing the high accuracy of the fluid flow PINN model.

\begin{table*}[htbp]
{\scriptsize
\centering
\caption{Hyperparameters for different PINN architectures and training settings in the 2D reaction-diffusion-advection problem.}
\renewcommand{\arraystretch}{1.3}
\begin{tabular}{@{}p{4.2cm}p{3.8cm}p{4.5cm}p{3.6cm}@{}}
\toprule
\textbf{PINN hyperparameter} & \textbf{Fluid Flow PINN} & \textbf{Mass Transport Sensitivity PINN} & \textbf{Mass Transport Inference PINN} \\
\midrule
Hidden layers & 8 & 5 & 5 \\
Neurons per layer & 60 & 100 & 100 \\
Inputs & 2 ($x$, $y$) & 5 ($x$, $y$, $t$, $Pe$, $Da$) & 3 ($x$, $y$, $t$) \\
Outputs & $\hat{v}$, $\hat{p}$) & $\hat{c}$ & $\hat{c}$ \\
Activation function & $\tanh$ & $\tanh$ & $\tan$ \\
Optimizer & Adam and L-BFGS & Adam and L-BFGS & L-BFGS \\
Optimizer iterations & 35000 (Adam) and 2000 (L-BFGS) & 35000 (Adam) and 1000 (L-BFGS) & 2000 \\
\addlinespace
Number of initially sampled training points & 
\begin{tabular}[t]{@{}l@{}}
$N_f = 40000 $ (collocation) \\
$N_{in} = 1500$ (inlet) \\
$N_{out} = 1500$ (outlet) \\
$N_{wall} = 500$ (channel wall) \\
$N_{obst} = 100$ (obstacle wall)
\end{tabular} &
\begin{tabular}[t]{@{}l@{}}
$N_f = 55000$ (collocation) \\
$N_{in} = 7000$ (inlet) \\
$N_{out} = 3000$ (outlet) \\
$N_{wall} = 4000$ (channel wall) \\
$N_{obst} = 3000$ (obstacle wall) \\
$N_{init} = 15000$ (initial condition)
\end{tabular} &
\begin{tabular}[t]{@{}l@{}}
$N_f = 75000$ (collocation) \\
$N_{in} = 8000$ (inlet) \\
$N_{out} = 8000$ (outlet) \\
$N_{wall} = 8000$ (channel wall) \\
$N_{obst} = 3000$ (obstacle wall) \\
$N_{init} = 15000$ (initial condition)
\end{tabular} \\
\addlinespace
Dynamic Weight Strategy & Yes & Yes & No \\
RAR-D & Yes & Yes & Yes \\
Frequency of sampling new training points &
Every 2000 (Adam) training iterations &
Every 2000 (Adam) training iterations &
Every 100 (L-BFGS) training iterations \\
Number of added training points with RAR-D &
\begin{tabular}[t]{@{}l@{}}
300 collocation points \\
20 inlet points \\
30 wall points 
\end{tabular} &
\begin{tabular}[t]{@{}l@{}}
1000 collocation points \\
300 inlet points \\
300 wall points \\
800 initial condition points
\end{tabular} &
\begin{tabular}[t]{@{}l@{}}
750 collocation points \\
50 inlet points \\
50 wall points \\
150 initial points
\end{tabular} \\
\bottomrule
\end{tabular}
\label{tab:pinn_hyperparams_all}
}
\end{table*}

\subsubsection*{Mass-Transport Sensitivity PINN}
We next construct a PINN model to compute the sensitivity of species concentration with respect to the dimensionless parameters $Pe$ and $Da$.
This model admits five inputs: $x$, $y$, $t$, $Pe$, and $Da$, and outputs $\hat{c}$.
Its architecture consists of 5 hidden layers with 100 neurons each and $\tanh$ activations.
The loss function is:
\begin{equation}
    \begin{split}
        \mathcal{L}(\theta) =\; & w_f\, \mathcal{L}_f(\theta) + w_{init}\, \mathcal{L}_{init}(\theta)
+ w_{in}\, \mathcal{L}_{in}(\theta)+ w_{out}\, \mathcal{L}_{out}(\theta)
 + w_{wall,up}\, \mathcal{L}_{wall,up}(\theta) \\
&+ w_{wall,low}\, \mathcal{L}_{wall,low}(\theta)
+ w_{obst}\, \mathcal{L}_{obst}(\theta)  + w_{\text{sens}, Pe}\, \mathcal{L}_{\text{sens}, Pe}
+ w_{\text{sens}, Da}\, \mathcal{L}_{\text{sens}, Da},
    \end{split}
\end{equation}

\noindent where $\mathcal{L}_f(\theta)$ is the residual of \Cref{eq: mass non_dim}, $\mathcal{L}_{init}(\theta)$ enforces the initial condition, $\mathcal{L}_{in}(\theta)$, $\mathcal{L}_{out}(\theta)$, $\mathcal{L}_{wall,up}(\theta)$, $\mathcal{L}_{wall,low}(\theta)$, $\mathcal{L}_{obst}(\theta)$ correspond to boundary condition residuals for the species concentration at the inlet, outlet and walls of the channel.
The sensitivity terms $\mathcal{L}_{\text{sens}, Pe}$, $\mathcal{L}_{\text{sens}, Da}$ are sums of the derivatives of all residuals with respect to $Pe$ and $Da$, respectively. 
Details are given in Section~II.2 of the Supplementary Material;
full training parameters are listed in \Cref{tab:pinn_hyperparams_all}. 
Since $Pe$ and $Da$ are initially unknown, this PINN model is trained using initial estimates (here, we use $Pe = 7.0$ and $Da = 18.0$).
As discussed in \Cref{sensitivity}, we assume that the maxima locations of the Gram determinant remain nearly unchanged when computed with these estimates instead of true values.
Training requires approximately 270 minutes.

\begin{figure}[ht]
    \centering
    \includegraphics[width=0.7\linewidth]{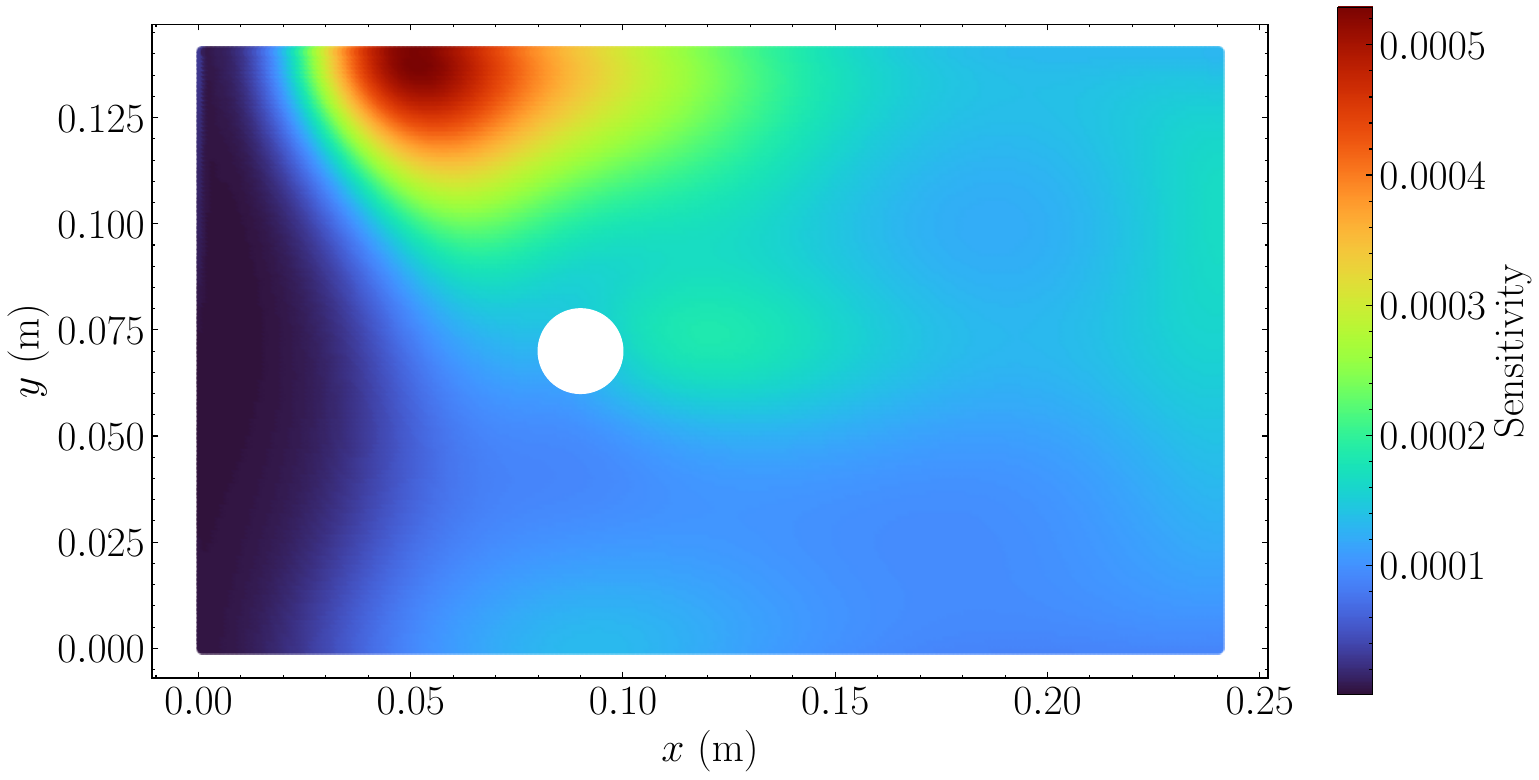}
    \caption{Spatial profile of Gram determinant computed with sensitivity PINN for a priori estimate of $Pe$ (7.0) and $Da$ (18.0)}
    \label{fig:apriori PINN gramdet}
\end{figure}

\begin{figure}[ht]
    \centering
    \includegraphics[width=0.6\linewidth]{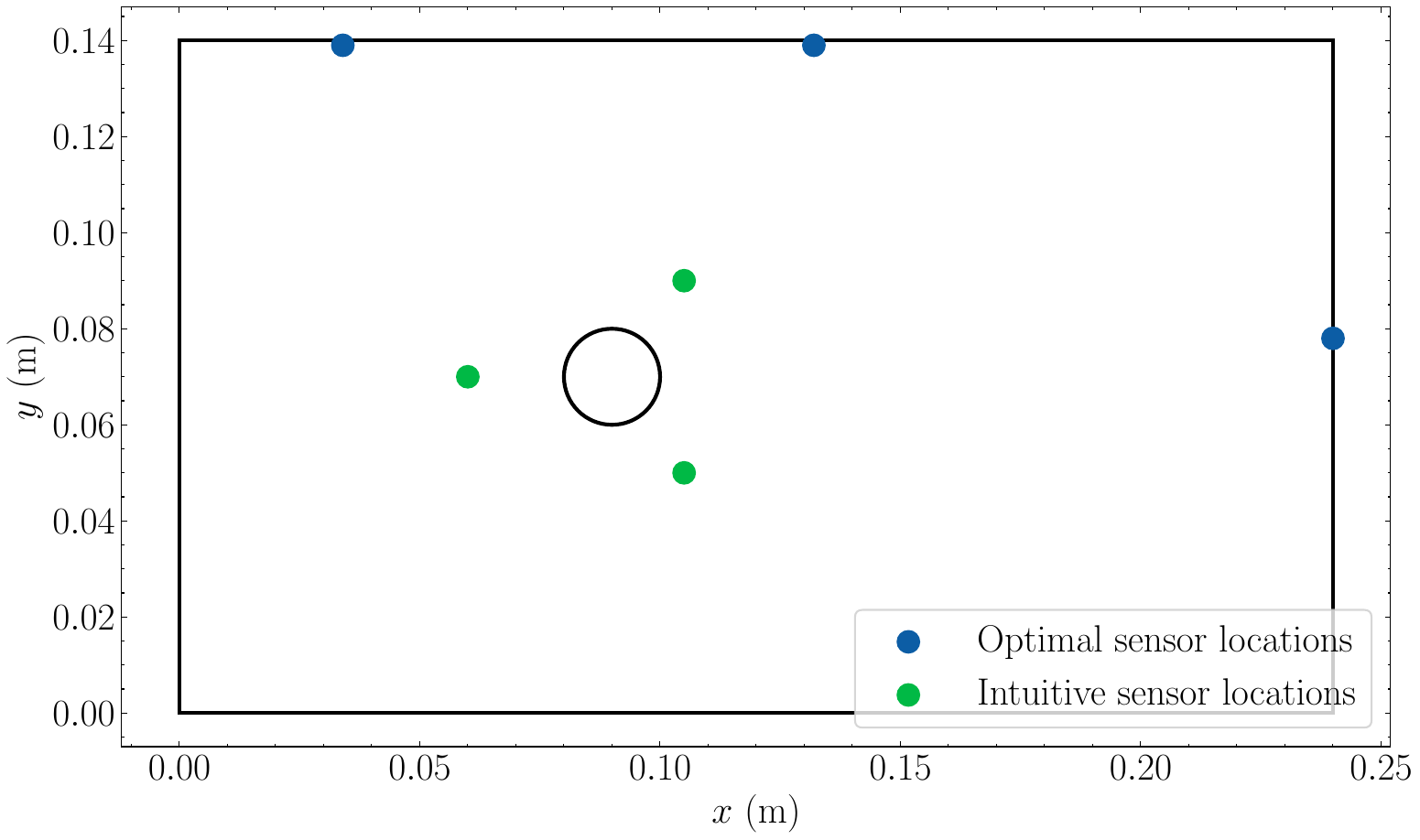}
    \caption{Optimally selected and intuitively placed sensor locations for the 2D set-up}
    \label{fig:random_vs_optimal_locations_2d}
\end{figure}

Once trained, the model provides sensitivity functions $s_{Pe}(x,y,t) = \frac{\partial \hat{c}}{\partial Pe}(x,y,t)$ and $s_{Da}(x,y,t) = \frac{\partial \hat{c}}{\partial Da}(x,y,t)$ computed via automatic differentiation.
These functions are used to assemble the FIM (\Cref{eq:M_vector,eq: FIM}), and evaluate its determinant for sensor placement.
For a single sensor, the Gram determinant depends only on $(x_1,y_1)$, and its spatial distribution is illustrated in \Cref{fig:apriori PINN gramdet}.
When multiple sensors are considered, the optimization of the Gram determinant becomes multidimensional.
To address this, we employ the Covariance Matrix Adaptation Evolution Strategy (CMA-ES) \citep{cma_paper}, a derivative-free algorithm well suited for high-dimensional, non-convex, and nonlinear problems (Python package $pycma$ \citep{cma_package}).
We first examine the case of three optimal sensor locations and compare the inference results against a configuration of three intuitively placed sensors around the obstacle (see \Cref{fig:random_vs_optimal_locations_2d}).
Pseudo-experimental data are generated by computational experiments using the finite element based commercial software COMSOL \textsuperscript{\textregistered}  setting $Pe=12.0$ and $Da=22.0$ (target values to be inferred by our PINN-based framework).
Virtual sensors record species concentration every 3 seconds. 

\subsubsection*{Mass-Transport PINN for Parameter Estimation}
\begin{figure*}[ht]
    \centering
    \includegraphics[width = 1.0\linewidth]{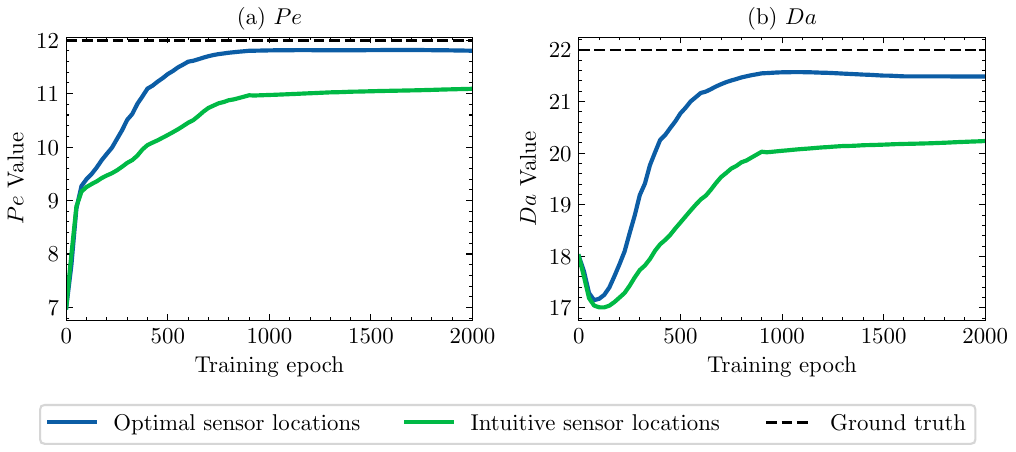}
    \caption{Parameter inference using data from optimally placed sensors versus intuitive placements: (a) $Pe$, (b) $Da$.}
    \label{fig: 2 sensor inference}
\end{figure*}

\begin{figure*}
    \centering
    \includegraphics[width=\linewidth]{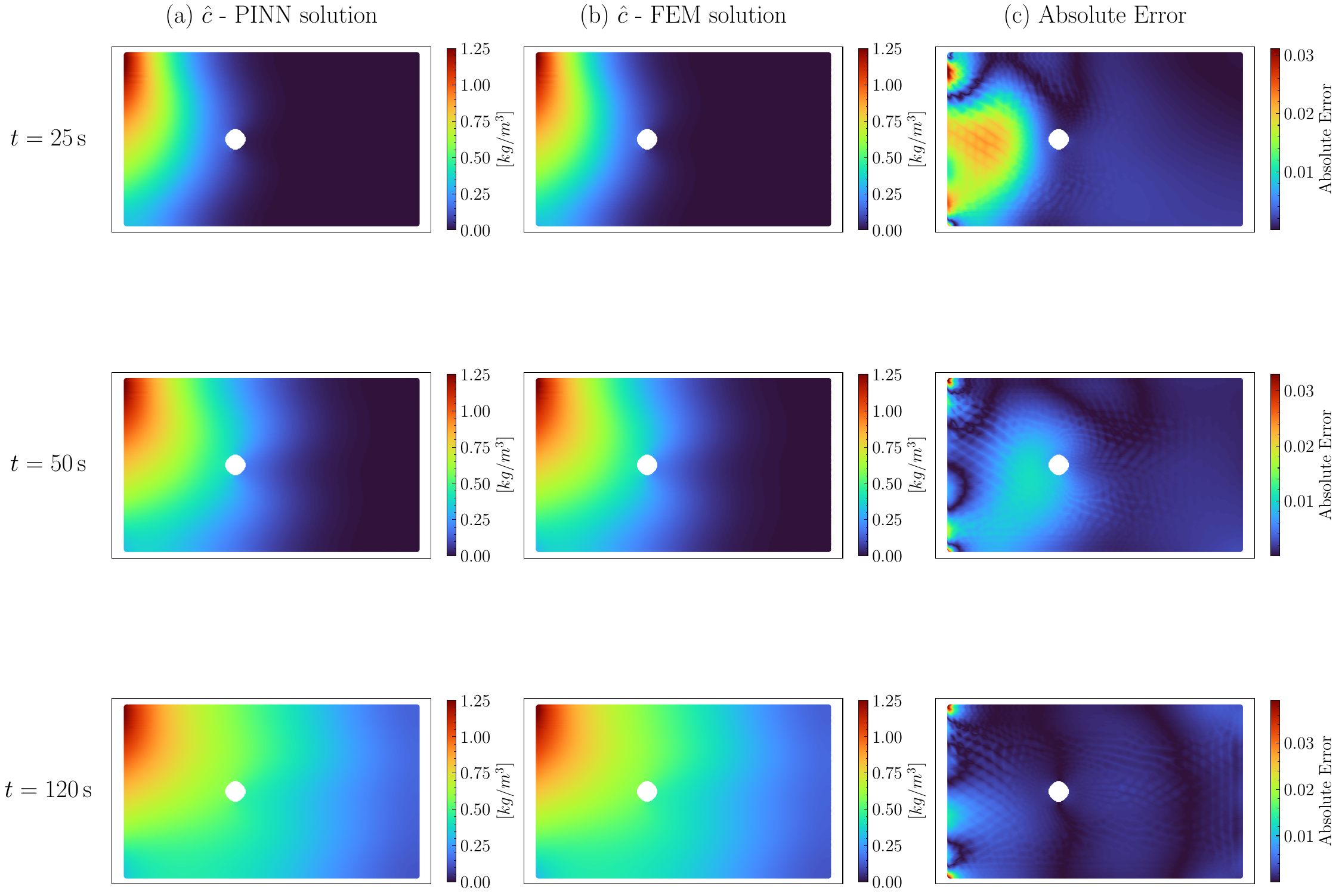}
    \caption{Predicted Concentration field $\hat{c}(x,y,t)$ at $t = 25, 50, 120$ s: (a) PINN prediction trained with data from optimal sensors, (b) FEM numerical solution for true parameter values, and (c) Absolute error ($|\hat{c}_{PINN} - \hat{c}_{FEM}|$).}
    \label{fig: concentration field}
\end{figure*}

Here, we train a PINN to infer the unknown parameters.
This neural network has the same architecture as the sensitivity PINN, but with only spatio-temporal inputs $(x,y,t)$.
Its loss function excludes the sensitivity terms ($\mathcal{L}_{\text{sens}, Pe}$ and $\mathcal{L}_{\text{sens}, Da}$) and instead incorporates a data loss term:
\begin{equation}
    \mathcal{L}_{data}(\mathbf{\theta}) = \frac{1}{N_{data}} \sum_{i=1}^{N_{data}} \left| \hat{c}({x}_{data}^{i}, y_{data}^i, t_{data}^i) - c_{data}^i \right|^2.
\end{equation}
For computational efficiency, we first pre-train this network on the forward problem with initial parameter estimates, then reuse this state for parameter estimation.
During fine-tuning, the weights and biases of the first three layers are frozen (see \Cref{sensitivity}).
The dynamic weight strategy  is disabled, and to emphasize observational data we set $w_{data} = 100$ (weight of data loss term), while keeping all other weights equal to 1.
Further training details are provided in \Cref{tab:pinn_hyperparams_all}.
The inference results for the three-sensor configurations are shown in \Cref{fig: 2 sensor inference}.
Both parameters are estimated with higher accuracy when using optimally placed sensors. 
In addition, the three optimally placed sensors provide sufficient information for the PINN to reconstruct the concentration field, as illustrated in \Cref{fig: concentration field}.

\begin{figure}[ht]
    \centering
    \includegraphics[width=0.95\linewidth]{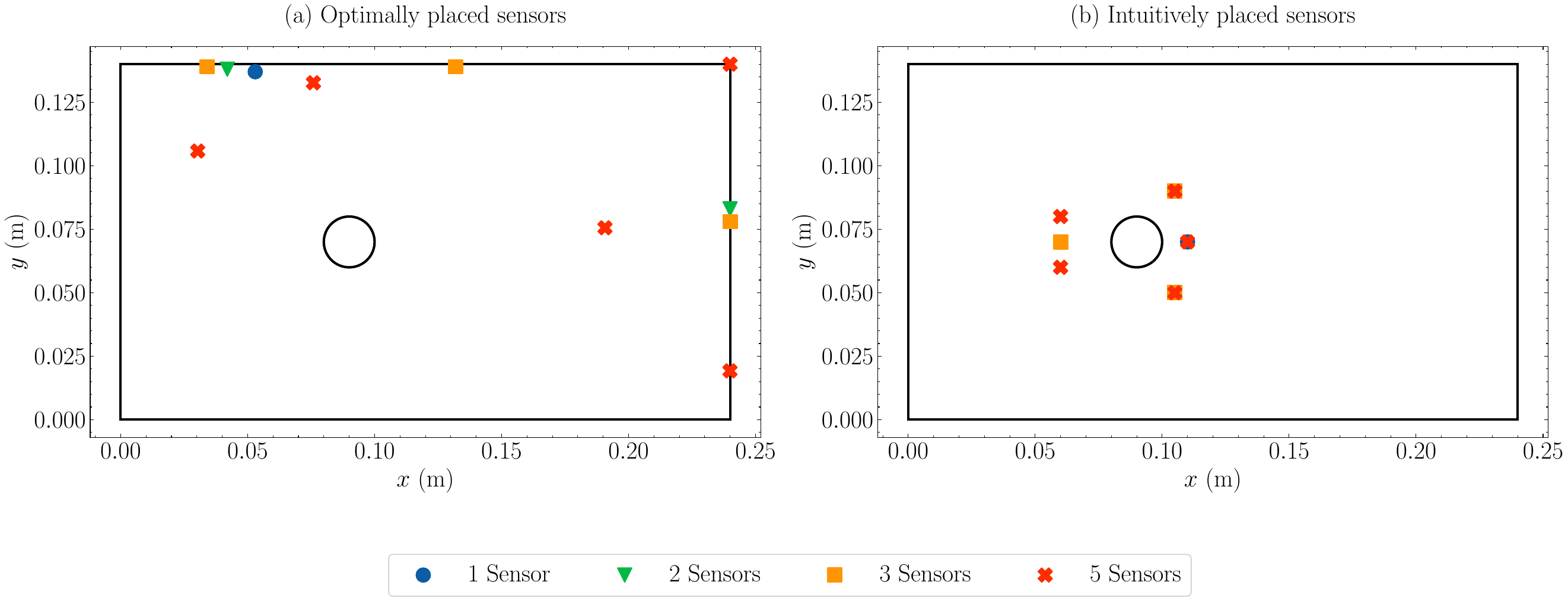}
    \caption{(a) Optimally selected and (b) intuitively placed sensor locations for the 2D set-up using 1 sensor (blue circle), 2 sensors (green triangles), 3 sensors (orange rectangles) and 5 sensors (red crosses).}
    \label{fig:random_vs_optimal_locations_2d_all}
\end{figure}

To study the effect of the number of sensors, we repeat the procedure for 1, 2 and 5 sensors, each time optimizing sensor placement using CMA-ES and comparing the results with intuitive arrangements.
For completeness, we show the locations of optimally placed sensors and the intuitively selected arrangements for 1,2,3 and 5 sensors in \Cref{fig:random_vs_optimal_locations_2d_all}.

\Cref{tab:inferred_params} summarizes the inferred parameter values, showing that optimally placed sensors consistently yield more accurate estimates.
Notably, adding more {\it intuitively} placed sensors does not necessarily improve performance, while accuracy increases monotonically with the number of {\it optimally} chosen sensors.

\begin{table}[ht]
\caption{Inferred $Pe$ and $Da$ values and relative error from ground truth using data (noise-free) for different sensor counts.}
\label{tab:inferred_params}
\centering
\begin{tabular}{@{}lcccc@{}}
\toprule
\textbf{Number of sensors} 
& \multicolumn{2}{c}{\textbf{$Pe$ (12.0)}} 
& \multicolumn{2}{c}{\textbf{$Da$ (22.0)}} \\
\cmidrule(lr){2-3} \cmidrule(lr){4-5}
 & Optimal sensors& Intuitive sensors& Optimal sensors& Intuitive sensors\\
\midrule
1 & 11.36 (5.33\%)& 11.09 (7.58\%)& 20.79 (5.50\%)& 20.33 (7.59\%)\\
2 & 11.62 (3.17\%)& 10.65 (11.25\%)& 21.08 (4.18\%)& 19.28 (12.36\%)\\
3& 11.80 (1.67\%)& 11.09 (7.58\%)& 21.49 (2.32\%)& 20.23 (8.05\%)\\
5& 11.98 (0.17\%)& 11.34 (5.50\%)& 21.78 (1.00\%)& 20.72 (5.82\%)\\
\bottomrule
\end{tabular}
\end{table}

We further tested robustness to noise adding Gaussian noise to the  COMSOL-generated pseudo-experimental  concentration data. 
Each measurement was drawn from:

\begin{equation}
    \tilde{c}^i_{data} \sim \mathcal{N}(c^i_{data}, (0.1c^i_{data})^2).
\end{equation}

\begin{figure*}[htbp ]
    \centering
    \includegraphics[width = 1.0\linewidth]{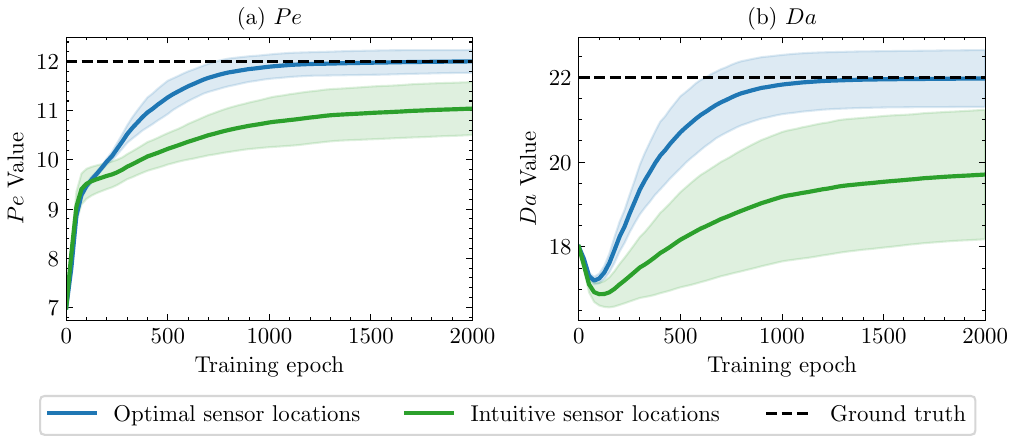}
    \caption{Mean and standard deviation (shaded area) of parameter inference with noisy data from optimal sensor locations versus intuitive sensor placements: (a) $Pe$, (b) $Da$}
    \label{fig: 3 noisy sensor inference}
\end{figure*}

\begin{figure}[htbp]
    \centering
    \includegraphics[width=1\linewidth]{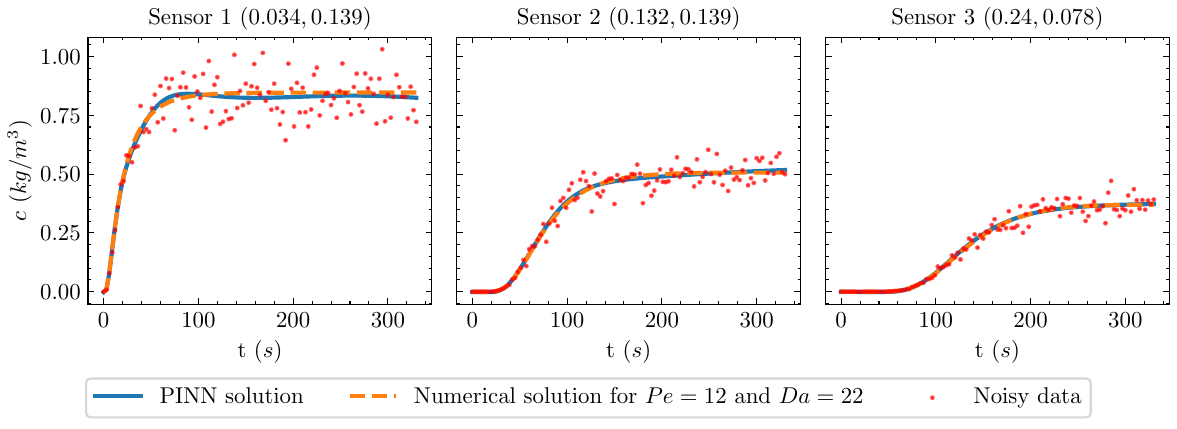}
    \caption{PINN prediction fitted to noisy data from 3 optimally placed sensors, compared with FEM solution at true parameters.}
    \label{fig: noisy data fit}
\end{figure}

\noindent To account for the noisy data, we repeated the parameter estimation training five times for each configuration.
\Cref{fig: 3 noisy sensor inference} depicts results for the three-sensor case, showing mean trajectories and standard deviations.
Optimal sensors again led to more accurate averages, particularly for $Da$, which corresponds to the nonlinear term of \Cref{eq: mass non_dim}.
The same trends hold for the 1-, 2-, and 5-sensor cases (\Cref{tab:inferred_params noisy}).
\Cref{fig: noisy data fit} compares the PINN solution (fitted to noisy data from 3 optimal sensors) with the FEM solution at the true parameters.
One can observe that the PINN predictions are in close agreement with the sensor data in terms of overall trends.
Each estimation experiment required approximately 100 minutes of training.

\begin{table}[htbp]
\caption{Mean Inferred $Pe$ and $Da$ values (with standard deviation) using noisy data for different sensor counts.}
\label{tab:inferred_params noisy}
\centering
\begin{tabular}{@{}lcccc@{}}
\toprule
\textbf{Number of sensors} 
& \multicolumn{2}{c}{\textbf{$Pe$ (12.0)}} 
& \multicolumn{2}{c}{\textbf{$Da$ (22.0)}} \\
\cmidrule(lr){2-3} \cmidrule(lr){4-5}
 & Optimal sensors& Intuitive sensors& Optimal sensors& Intuitive sensors\\
\midrule
1 & $10.73 \pm 0.61$& $10.73 \pm 0.19$& $20.30 \pm 1.14$  & $19.47 \pm 1.02$\\
2 & $11.38 \pm 0.35$& $11.38 \pm 0.76$  & $20.20 \pm 0.97$& $20.01 \pm 1.07$
\\
3& $12.01 \pm 0.26$& $11.18 \pm 0.60$& $21.97 \pm 0.75$& $19.59 \pm 1.94$\\
5& $11.96 \pm 0.43$& $11.12 \pm 0.38$& $21.26 \pm 0.55$& $20.11 \pm 0.97$\\
\bottomrule
\end{tabular}
\end{table}

\section{Conclusions}\label{conclusions}

In this work, we develop a novel ML-based methodology for optimal sensor placement in large-scale distributed parameter systems for efficient parameter estimation. We  integrate the D-optimal criterion for sensor placement into the PINN framework, with the goal of leveraging PINNs' parameter estimation capabilities using minimal observational data. 
%
%
Our approach first determines optimal sensor locations based on an {\it a priori} estimate for the parameters of interest, and then identifies these parameters using data collected from those locations.
To solve the optimal sensor placement problem, we train a PINN that takes as input both the parameters of interest and the spatio-temporal coordinates. 
This design enables the computation of sensitivity functions via automatic differentiation, which are then used to select the optimal sensor positions. 
For parameter inference, we employ a standard PINN architecture trained on data sampled from either optimally placed or intuitively chosen virtual sensors generated from numerical solutions.
We evaluate the framework on two illustrative case studies: A 1D steady-state and a 2D transient reaction-diffusion-advection problem.
In the 1D case, we estimate the P{\'e}clet number using a single sensor.
Placement at the optimal location yields a relative error of only {0.20\%}, whereas placement at the outlet produces a substantially larger error of {84.7\%}, underscoring the importance of optimal sensor positioning.
In the 2D transient case, we jointly estimate the P{\'e}clet and Damk{\"o}hler numbers using sensor configurations of 1, 2, 3, and 5 sensors.
With noiseless data, optimal sensor placements consistently provide more accurate estimates, and accuracy improves as the number of sensors increases. 
Introducing measurement noise preserved these trends.
Training a PINN to compute sensitivity functions is computationally intensive at the outset ($\approx 270$ minutes), owing to the additional loss terms requiring derivatives of all residuals with respect to all parameters of interest.
However, once trained, transfer learning enables rapid re-training, offering substantial flexibility.
This allows the framework to adapt efficiently to changes in boundary or initial conditions, modifications to system geometry, or variations in the underlying physics (e.g., altered reaction kinetics), without restarting training from scratch. 
Moreover, since PINNs are mesh-less, sensitivity functions can be computed at any spatio-temporal location.

Overall, our findings demonstrate that PINNs can be effectively used to compute sensitivity functions for optimal sensor placement in parameter estimation.
Data collected at these locations significantly improve estimation accuracy, even under measurement noise.
Beyond the benchmark problems considered here, the framework can be extended to more complex systems with multiple unknown parameters or partially defined models (e.g., unknown kinetics in plasma reactors), where data acquisition is costly or limited.

\section*{Acknowledgements}

CT acknowledges funding through the program Y$\Pi$2TA-0559703 implemented within the framework of the National Recovery and Resilience Plan "Greece 2.0" and financed by the European Union (NextGeneration EU)





\clearpage 






\printcredits

\bibliographystyle{cas-model2-names}

\bibliography{cas-refs}



\end{document}